\def\year{2022}\relax

\documentclass[letterpaper]{article} 
\usepackage{aaai23}  
\usepackage{times}  
\usepackage{helvet}  
\usepackage{courier}  
\usepackage[hyphens]{url}  
\usepackage{graphicx} 
\usepackage{natbib}   
\usepackage{caption}  
\usepackage{algorithm}
\usepackage{algorithmic}
\usepackage{amssymb}

\usepackage{newfloat}
\usepackage{listings}
\floatstyle{ruled}
\newfloat{listing}{tb}{lst}{}
\floatname{listing}{Listing}

\title{Forecasting with Sparse but Informative Variables:\\ A Case Study in Predicting Blood Glucose}
\author{
    Harry Rubin-Falcone\textsuperscript{\rm 1}, Joyce Lee\textsuperscript{\rm 2}, Jenna Wiens\textsuperscript{\rm 1}
}
\affiliations{
    \textsuperscript{\rm 1}Division of Computer Science and Engineering, University of Michigan\\
    \textsuperscript{\rm 2}Division of Pediatric Endocrinology, University of Michigan\\
}

\usepackage{bibentry}

\begin{document}

\maketitle

\begin{abstract}
In time-series forecasting, future target values may be affected by both intrinsic and extrinsic effects. When forecasting blood glucose, for example, intrinsic effects can be inferred from the history of the target signal alone (\textit{i.e.} blood glucose), but accurately modeling the impact of extrinsic effects requires auxiliary signals, like the amount of carbohydrates ingested. Standard forecasting techniques often assume that extrinsic and intrinsic effects vary at similar rates. However, when auxiliary signals are generated at a much lower frequency than the target variable (\textit{e.g.}, blood glucose measurements are made every 5 minutes, while meals occur once every few hours), even well-known extrinsic effects (\textit{e.g.}, carbohydrates increase blood glucose) may prove difficult to learn. To better utilize these \textit{sparse but informative variables} (SIVs), we introduce a novel encoder/decoder forecasting approach that accurately learns the per-timepoint effect of the SIV, by (i) isolating it from intrinsic effects and (ii) restricting its learned effect based on domain knowledge. On a simulated dataset pertaining to the task of blood glucose forecasting, when the SIV is accurately recorded our approach outperforms baseline approaches in terms of rMSE (13.07 [95\% CI: 11.77,14.16] vs. 14.14 [12.69,15.27]). In the presence of a corrupted SIV, the proposed approach can still result in lower error compared to the baseline but the advantage is reduced as noise increases. By isolating their effects and incorporating domain knowledge, our approach makes it possible to better utilize SIVs in forecasting. 
\end{abstract}
\section{Introduction}

In time-series forecasting, the future values of a target signal can depend on both intrinsic and extrinsic effects. Intrinsic effects are dynamics that depend only on the current and past values of the target signal. In contrast, extrinsic effects are dynamics that arise due to auxiliary variables. In many cases, the inclusion of such auxiliary signals as input to a forecasting model, in addition to the target signal, results in more accurate forecasts \cite{multiv1,book}. However, in other settings, including auxiliary variables as input to a forecasting model produces little to no improvement in forecast accuracy, even when there is a known relationship between the additional variables and the target signal. This is particularly true in forecasting physiological variables like blood glucose. Auxiliary signals like carbohydrates consumed and bolus insulin administered both have well-known effects on blood glucose, but their inclusion as inputs to forecasting models has not, in general, led to significant improvements in performance over models based on blood glucose alone \cite{drf,klein,marshall}. We hypothesize that this is due in part to a mismatch in the relative frequency of non-zero values between the auxiliary signal and the target signal. We refer to forecasting tasks where an auxiliary signal is sparse but has a known effect on the target signal as the \textit{sparse but informative variable (SIV)} problem. A forecasting model that successfully addresses this problem will leverage the SIV despite its sparsity, leading to overall improved predictions.

\begin{figure}[t]
  \centering 
\hspace*{-.4cm}  \includegraphics[height=1.4 in]{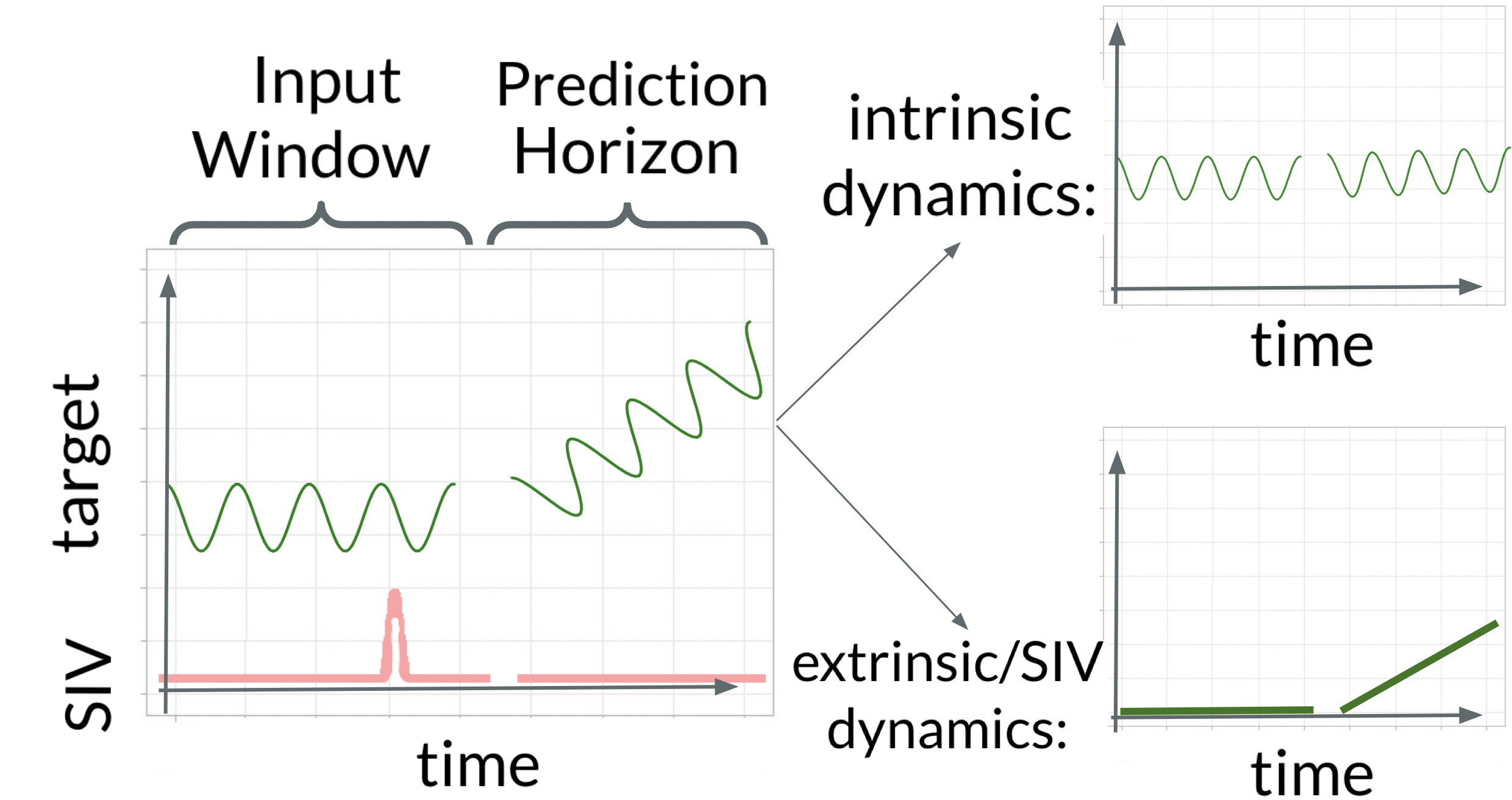} 
  \caption{An overview of the SIV problem. In this toy example, the target variable exhibits oscillatory behavior when only zero SIV values are present (intrinsic dynamics), and the presence of a non-zero SIV value causes the target signal to increase linearly (extrinsic [SIV] dynamics). }\label{fig:overview}
\end{figure}

\textbf{Problem Definition.} In this work, we introduce and address the SIV problem (\textbf{Figure \ref{fig:overview}}), which arises when an auxiliary variable that occurs infrequently is known to cause an increase or decrease in the target variable's magnitude over time, although the exact effect may be unknown. The sparsity of the SIV often results in the failure of standard multi-input forecasting approaches in leveraging the auxiliary variable, \textit{i.e.}, models that include the variable perform similarly to univariate-input approaches. A model that has overcome the SIV problem utilizes the SIV in making its predictions, resulting in improved forecasting accuracy relative to a model that does not use the additional variable. The SIV problem occurs when an important variable is mostly zero-valued. This is \textit{not} the same as a sparsely \textit{sampled} variable (SSV). In the case of under-sampling, the variable is sparse because it is not measured. In the SIV problem, we assume that the variable is measured frequently and accurately, but for most timepoints it is zero. We examine our setting as this assumption is relaxed and noise is added, but we consider SIVs to be generally noise-free, as noise and missingness pose a separate problem. While approaches for addressing missingness or SSVs have been extensively studied \cite{missing}, the SIV problem has not.

\textbf{Challenges.} Although developing forecasting strategies that  make use of SIVs has the potential to improve predictive accuracy in several domains, it has not been directly addressed in previous work.  Recent multivariate forecasting approaches attempt to learn complex inter-variable dependencies \cite{attention1,cnn2,sharedmulti}. However, these approaches do not explicitly account for the relative sparsity of some variables. Naive approaches to addressing the SIV issue include resampling the training data to include more samples with non-zero SIV values and carrying forward the SIV values to the end of the input window, but in practice we have found that these approaches generally fail to improve performance.  As we will demonstrate, the incorporation of domain knowledge in terms of restricting model outputs could help encourage a forecasting model to make better use of the SIV. However, existing state-of-the-art deep forecasting approaches generally do not use such restrictions in order to maintain flexibility. Here, we strive for a combination of the two: a forecasting approach that maintains flexibility while incorporating domain knowledge.

\textbf{Our Idea.} To address the SIV problem we propose a novel forecasting approach: ``The Linked Encoder/Decoder.''  Our model integrates two main ideas: (i) the isolation of intrinsic and extrinsic effects, and (ii) the incorporation of domain knowledge. We implement the first idea with two separate but connected decoder networks. One network learns per-timepoint SIV effects (the SIV network), and the other learns the intrinsic dynamics of the target variable (the target network). We implement the second idea by restricting the output of the SIV network based on domain knowledge. Combined, these ideas lead to overall improved usage of the SIV and in turn more accurate forecasts.

Our main contributions are as follows:

• We present the sparse informative variable (SIV) problem.

• We propose a novel forecasting approach designed to  leverage the SIV by isolating the effect of the SIV and incorporating domain knowledge.

• We evaluate our model in the context of forecasting blood glucose measurements and show that it more effectively incorporates SIVs compared to several baselines, even in the presence of a small amount of noise.

\section{Problem Setup}
Here, we formalize our task and describe our motivating setting: blood glucose forecasting in type 1 diabetes.

\textbf{Task Formalization.} We focus on the task of multi-input univariate-output time-series forecasting in which we aim to predict the future values of a single target variable $x\in\mathbb{R}$, but have access to an additional auxiliary variable $x'\in\mathbb{R}$ that is sparse but informative. More specifically, $x'$ is zero at a much higher frequency than the target signal and the presence of a non-zero $x'$ value has a known effect on the target signal (\textit{e.g.}, it results in either an increase or decrease). Given data pertaining to the previous $T$ values of the target signal, $\mathbf{x}_{-T+1:0}$, and the auxiliary signal $\mathbf{x}'_{-T+1:0}$, we aim to predict the next $h$ timepoints of the target signal: $\mathbf{y}=\mathbf{x}_{1:h}$.

As is common in forecasting work \cite{bbook}, we assume the target signal is generated by some underlying yet unknown autoregressive process. We assume  non-zero SIV values contribute in an additive autoregressive way.  Our setup focuses on the setting in which extrinsic effects are driven by an SIV, but there are settings where extrinsic effects may be driven by a variable that is not sparse, or a combination of sparse and non-sparse variables.  We focus on the SIV problem and assume that any additional non-sparse extrinsic effects can be modelled using standard forecasting approaches.

\textbf{A Motivating Example- Predicting Blood Glucose.} The SIV problem arises in blood glucose (BG) forecasting, which has been extensively studied \cite{glureview}, including in deep learning settings \cite{drf,iankdd,gluexps}. In BG forecasting, one aims to estimate BG concentration for some prediction horizon, based on a history of BG measurements and other input signals. This is a challenging task because glucose dynamics vary based on activity, time of day, hormone levels and more, resulting in significant non-stationarity.  Individuals with type 1 diabetes are unable to produce insulin, which is critical for maintaining healthy BG levels. They therefore need to manually administer insulin to manage BG levels, which requires constant monitoring and frequent decisions regarding timing and dosage. Improved accuracy of BG forecasts enables timely and accurate treatment and therefore improves BG management.

Consuming carbohydrates when eating a meal increases BG, while insulin decreases BG. Insulin boluses and carbohydrates are considered SIVs, since they occur only a few times a day. In contrast, BG is recorded every five minutes as unique, non-zero values. Carbohydrates and insulin alter BG after a delay of 30 minutes to an hour. The effects of these variables are transient, so BG forecasters can learn to ignore these variables while still generating accurate predictions for most timepoints. In fact, current forecasting approaches do as well without information on carbohydrates and insulin as with \cite{drf,klein,marshall}.  However, accurate forecasts following insulin and carbohydrate measurements are particularly important because those are the time periods with the greatest BG variability and therefore present the greatest risk to an individual.

\section{Methods}

\begin{figure*}[t]
  \centering 
r\hspace*{-.4 cm}  \includegraphics[height=2 in]{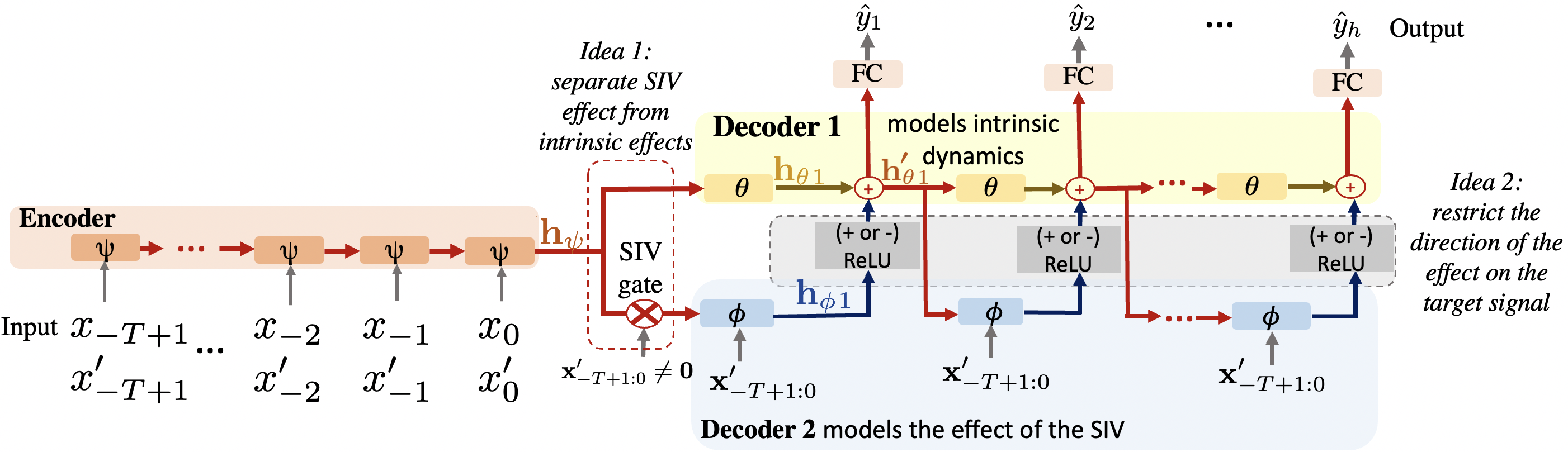} 
  \caption{Our architecture: the Linked Encoder/Decoder, shown with an input length $T$ and prediction horizon $h$. The $\theta$ network models intrinsic dynamics, while the $\phi$ network models the SIV dynamics. The $\psi$ network models shared dynamics. Input time-series are gated, such that only inputs ($[\mathbf{x}_{-T+1:0},\mathbf{x}'_{-T+1:0}]$) containing a non-zero SIV at any timepoint are passed through the $\phi$ network. The ReLU network shown in grey is used to ensure that the relationship between SIV and target is as expected.  }\label{fig:model2}
\end{figure*}

  \textbf{Overview.} To effectively capture the autoregressive dynamics of forecasting  with an SIV, our architecture, the ``Linked Encoder/ Decoder,''  relies on a recursive framework (\textbf{Figure \ref{fig:model2}}). It involves one encoder network and two linked decoder networks, which are used to isolate the SIV dynamics from the intrinsic dynamics. The SIV decoder receives the SIV signal as input. The decoder systems share a hidden state, which is processed in parallel. This allows the intrinsic and extrinsic dynamics to be learned separately. Once isolated, we restrict the effect of the SIV on the target signal based on domain knowledge.

\textbf{Standard Encoder/Decoder.} Our approach is based on a standard encoder/decoder recurrent neural network, as depicted by the orange and yellow sections of \textbf{Figure \ref{fig:model2}}. For samples where $\mathbf{x}_{-T+1:0}'=\mathbf{0}$, this encoder/decoder is not modified. A single encoder ($\psi$),  takes  $\mathbf{x}_{-T+1:0}$ and  $\mathbf{x}'_{-T+1:0}$ as input. The encoder outputs a hidden state $\mathbf{h}_{\psi}$= $\psi([\mathbf{x}_{-T+1:0} ; \mathbf{x}'_{-T+1:0}])$, which is passed through a decoder LSTM ($\theta$) that outputs hidden state $\mathbf{h}_{{\theta}1}=\theta(\mathbf{h}_{\psi})$. At each timepoint in the forecast horizon, the output from the previous time step $\mathbf{h}_{{\theta}t-1}$ is passed through $\theta$, such that $\mathbf{h}_{{\theta}t}=\theta(\mathbf{h}_{{\theta}t-1})$. This learned representation is also passed through fully connected output network $FC$ at each time step $t$ in the forecast horizon to output a prediction $\hat{y}_t=FC(\mathbf{h}_{{\theta}t})$.

\textbf{Linked SIV Decoder and Gating.} If an input sample contains a non-zero SIV value, it is passed through both this standard encoder/decoder and a second decoder $\phi$, depicted in the blue section of \textbf{Figure \ref{fig:model2}}. This second decoder aims to model SIV dynamics. By gating samples based on SIV values, we separate the extrinsic effects of the SIV on the target variable from the intrinsic effects of the target signal on itself. When the corresponding SIV values are non-zero (\textit{i.e.}, $\mathbf{x}_{-T+1:0}'\neq \mathbf{0}$), the network engages the second decoder, which processes hidden state $\mathbf{h}_{{\theta}t}$ for $t=1,...,h$, in parallel with $\theta$, as described below. Because $\phi$ is only engaged when an SIV is present, $\theta$ learns to forecast in the absence of an SIV, while $\phi$ learns the additional effect of the SIV.

 \textbf{Linked Hidden State Processing and Incorporating Domain Knowledge.} Both decoders process a single hidden state in parallel, and their outputs are summed. At the first time step in the prediction horizon, both decoders take as input $\mathbf{h}_{\psi}$, but they each output a unique hidden state ($\mathbf{h}_{{\theta}t}$ and $\mathbf{h}_{{\phi}t}$).  At subsequent time steps, these two hidden states are summed to create a new hidden state: as in Figure \ref{fig:model2}, we define $\mathbf{h}'_{{\theta}t} = \mathbf{h}_{{\theta}t} + k\cdot max(\mathbf{h}_{{\phi}t},\mathbf{0})$, where $k=1$ if the SIV is known to lead to an increase in the target signal and $k=-1$ otherwise. This is equivilent to passing $\mathbf{h}_{{\phi}t}$ through a ReLU function, which restricts the effect of the SIV on the target variable to the expected direction. The combined hidden state ($\mathbf{h}'_{{\theta}t}$) is passed to the output network ($FC$) and both decoders at subsequent time steps.  The final forecast $\hat{\mathbf{y}}$ is a product of both decoders, capturing both intrinsic and restricted extrinsic effects (\textit{i.e.}, $\hat{y}_t=FC(\mathbf{h}_{{\theta}t}')$). Note that when $\mathbf{x}'=\mathbf{0}$, $\phi$ is not engaged, and $\mathbf{h}'_{{\theta}t}=\mathbf{h}_{{\theta}t}$.
 
In order to encourage the SIV decoder ($\phi$) to utilize the SIV, $\phi$ receives the entire SIV signal as input, concatenated with the hidden state. The SIV signal is shifted at each timepoint so that the encoder's position in time relative to the SIV is included in the representation implicitly (see implementation details).

\textbf{Additional Variables.} In \textbf{Figure \ref{fig:model2}}, we present an overview of the proposed architecture for a setting with a single SIV. In a setting with multiple SIVs, the number of secondary decoders is increased and restrictions are applied according to the known effect of each SIV. Each $\phi$ takes as input only the relevant SIV signal, along with $\mathbf{h}_{{\theta}t}'$.  $\mathbf{h}_{{\theta}t}'$ is modified by all SIV decoder systems, so that the hidden state that is passed to $FC$ and subsequent decoder steps is a sum of the number of SIVs plus one components. Non-sparse auxiliary variables, if any, are given to the $\psi$ network, along with $\mathbf{x}$ and $\mathbf{x}'$, so that non-SIV extrinsic effects can be modeled (by $\theta$, since these variables do not effect gating). 

\begin{figure}[t]
  \centering 
\hspace*{-.5cm}  \includegraphics[height=1.8 in]{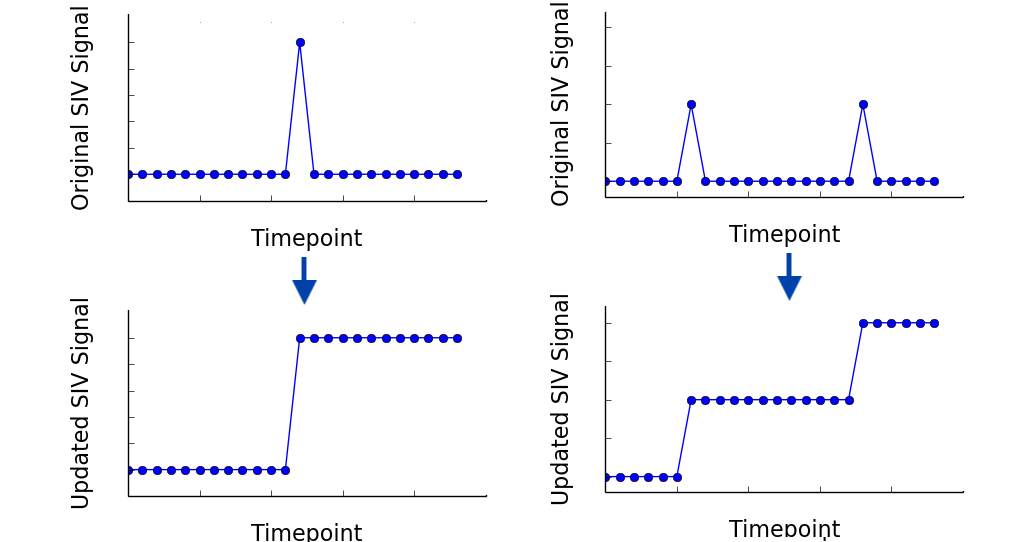}
\caption{Our sum-total approach. We use the sum total up to the current point within an input window as input. This method allows the SIV signal to make a larger impact on the gradient while maintaining all temporal information. }\label{fig:cf}
\end{figure}

\textbf{SIV Representation.} One issue that makes utilizing SIVs difficult is that they usually occur at only one timepoint in an input window, having little effect on the gradient during training.  To increase its effect, we use the sum-total SIV value up to the current timepoint as input (\textbf{Figure \ref{fig:cf}}). Up until the first non-zero SIV value of an input time-series, the signal value is zero. After any non-zero SIV, the input is the sum of observed values prior to and including that point, \textit{in the input time-series only}. SIV values from before the input window are ignored. This approach is used for all analyses, including baselines and ablations.  This sum-total input improved performance for all approaches (see \textbf{Appendix B}).

\section{Experimental Setup}

We evaluate our approach on several datasets pertaining to BG forecasting in type 1 diabetes. We compare to several baselines, including resampling approaches. We evaluate forecast accuracy and the extent to which each model utilizes the SIVs.

\subsection{Forecasting Task}
 
 We aim to forecast BG 30 minutes into the future ($h=6$), based on a history of BG and two SIVs: carbohydrate and insulin bolus values. We use an input length of 2 hours ($T=24$). Here, $h=6$ as it represents a common BG forecasting benchmark \cite{kn:oc} and we set T=24 based on prior work that suggests longer histories do not provide additional benefit \cite{glureview}.  Target and auxiliary variables are scaled to between zero and one, using the maximum expected value to linearly scale (400 for BG, 200 for carbs, 50 for insulin). Each individual's train/validation/test data is split into overlapping windows of length $T+h$ with a stride of 1, to be used as model input and labels.  

\subsection{Data}

We compare the performance of our architecture to baselines on three type 1 diabetes-based datasets: Simulated, Simulated-noisy, and Ohio. All three datasets are publicly available \cite{uva,simgit,kn:oc}.

\textit{Simulated.} Data generated from a commonly-used type 1 diabetes simulator provide a curated test setting on which to evaluate our approach. We use the UVA-Padova simulator \cite{uva}, via a publicly available implementation \cite{simgit}. We generate ten days of data for ten individuals (the ten ``adult'' patients modeled in the simulator), corresponding to 28,800 timepoints. Carbohydrate and insulin values occur every 111 timepoints on average (carbohydrates median, [IQR]: 84 timepoints between occurrences, [60,150]. boluses: 83, [58,148]). The meal schedule used to generate simulated data is based on the Harrison-Benedict equation \cite{meal} as implemented in \cite{deepblood}, but without snacks (3 meals a day), to further highlight the SIV problem.  We use the default basal-bolus controller from the existing implementation to administer insulin, but delay three quarters (randomly selected) of the bolus administrations to be 20 minutes to two hours after the simulated meal, randomly sampled from a uniform distribution. This delay was to control for a separate issue that can arise when multiple SIVs occur at precisely the same time. The delay disentangles their effects.

\textit{Simulated - noisy.} While the simulator used above introduces noise in the BG measurements, we assume that the SIV signals are recorded without noise or missingness. To measure the robustness of our approach to this assumption, we generate additional simulated datasets using the approach above in which we vary the amount of noise and missingness. After data generation, we zero out between 10\% and 50\% of the carbohydrate values. Separately, we also explore the effects of adding uniform random noise to the measurements, varying the maximum magnitude between 10\% and 50\%. These changes are made to both training and testing data.

\textit{Ohio.} Finally, we examine performance on a real dataset that was made publicly available for the Knowledge Discovery in Healthcare Data BG Level Predication Challenge \cite{kn:oc}.
The data pertain to 12 individuals with BG measurements every 5 minutes. Carbohydrate values occur every 88 timepoints on average (median, [IQR]: 70, [56,134]). Boluses occur every 52 timepoints on average (median, [IQR]: 36, [28,63]). Carbohydrate measurements are input by the individual and as a result are subject to noise and missingness. However, unlike the `Simulated - noisy' dataset, we can neither measure nor control the extent of the noise in our experiments. Further details are provided in \textbf{Appendix A}.

\subsection{Baselines}

The carry-forward approach is used to represent SIVs for all baselines.

\textbf{Encoder/Decoder} (Abbrev: Enc/Dec). Our primary baseline is a stand-alone encoder/ decoder system, identical to the $\psi$ plus $\theta$ setup in our full architecture \cite{iankdd}. This baseline has less capacity than our proposed approach, so we increase the minimum number of training iterations to match the same number of gradient updates for our approach. We also examine a model with capacity that matches our proposed approach (Full Capacity).

\textbf{Full Capacity.} To ensure that any performance improvements observed are not due to our model's increased capacity, we also compare  to a model based on our full architecture, but with no SIV-specialization (\textit{i.e.}, there is no gating,  no direct SIV input into $\phi$, and no output restriction). This model is trained for the same number of iterations as our proposed approach.

\textbf{Resampling.} Resampling is perhaps the simplest approach to addressing signal sparsity. In order to rule out resampling as a naive solution to the SIV problem, we implement our primary baseline method with two resampling procedures: training the model on only windows with SIV samples to initialize the weights before training on the full sample (\textbf{SIV Initialize}), and training on the full sample, then fine-tuning the model on only windows with non-zero SIV values (\textbf{SIV Fine-tune}).  Similar to our primary encoder/decoder baseline, we increase the minimum number of training iterations to match the number of gradient updates used in training our proposed method.

\subsection{Implementation and Training Details}

\textbf{Implementation.} Each LSTM encoder or decoder is implemented as a 2-layer bidirectional LSTM with 100 hidden units. $FC$ is a fully connected linear network with a single output. Our architecture uses two $\phi$ networks, one for carbohydrates (positive effect, ReLU restriction) and one for bolus insulin (negative effect, $-1 \times $ ReLU restriction).  In order to input each SIV signal into each $\phi$ network while maintaining time information, $\mathbf{x}'_{-T+1:0}$ is front-padded with $h$ zeros and input to $\phi$ at the first time step of the forecast horizon. The signal is shifted back at each timepoint, such that at the $i^{th}$ time step of the prediction horizon, the SIV signal is shifted back $i-1$ positions, so that it is front-padded with $h-(i-1)$ zeros and back-padded with $i-1$ zeros. In this way the input corresponds with the encoder's position in time. $\mathbf{x}'_{-T+1:0}$ is scaled to have the same mean as $\mathbf{h}_{{\theta}t}'$ for each input.

\textbf{Training.} We split each individual's data into training, validation and test sets. We split data by number of timepoints using a 70\%/15\%/15\% split, without overlap. We implement and train our models in pytorch 1.9.1 and CUDA version 10.2, using Ubuntu 16.04.7 and a GeForce RTX 2080, using an Adam optimizer \cite{kn:adam} and a batch size of 500. We use a learning rate of 0.01 and a weight decay of $10^{-7}$. When training, mean square error (MSE) across all timepoints in the prediction horizon is used as a loss function. We train for at least 500 epochs, until validation performance does not improve for 50 epochs. The model parameters that led to the best validation set performance are used at inference time. We train and test a model on each individual and report across-individual averages.
 
\textbf{Reproducibility.} Our code, appendix, and simulated data are here: github.com/MLD3/sparse-informative-variables. The Ohio dataset can be made available through a data-use agreement (see \textbf{Appendix A}).

\subsection{Evaluation}

All evaluations are performed on held-out test sets. We measure forecasting performance using rMSE and mean absolute error (MAE). In order to match common practice in the BG forecasting literature, we calculate error terms based on the prediction accuracy of the final timepoint in the prediction window \cite{kn:oc}. Boot strap confidence intervals and test-statistics are calculated within each individual and then averaged.

\textbf{SIV Usage Metric.} Let $\mathbb{X}$ denote a dataset with an SIV, and let $\mathbb{X_\emptyset}$ denote the exact same dataset with all SIV values set to zero. Let $f$ define a mapping $f:X\rightarrow \mathbf{\hat{y}}$, where $X \in \mathbb{X}$, and $\mathbf{\hat{y}} \in \mathbb{R}^h$ is a prediction of the next $h$ points of the target variable. $f$ is trained and evaluated on $\mathbb{X}$, while $f_\emptyset$ is trained and evaluated identically to $f$, except using $\mathbb{X_\emptyset}$. Let $L$ denote the error of the model's prediction (here, rMSE or MAE). We define SIV usage as $L(f_\emptyset(\mathbb{X_\emptyset}))$ - $L(f(\mathbb{X}))$. It is inspired by the Shapley Regression Value \cite{shap}. This metric reflects how error changes when the SIV is removed. When removing the SIV, we both train and test on data without the SIV (rather than performing a permutation test or similar), so the model can learn the maximum amount of information available from the target variable alone.

\textbf{Individual-Level Analyses.} We compare the error and SIV usage of the baseline encoder/decoder to the improvement over baseline offered by our method across individuals.  We expect that our model will offer greater improvement over baseline for individuals with high baseline error and low baseline SIV usage, as those are the individuals for which SIVs are most poorly modeled in the baseline. This would indicate that our approach addresses baseline deficits in SIV-modelling.
 
\textbf{Ablations.}  We perform the following ablation analyses to examine which elements contribute to our model's performance. \textbf{No Gating}: All samples are passed through both decoders. \textbf{No Restriction}: The outputs of the SIV decoder systems are not passed through a ReLU function. \textbf{No SIV Input}: The SIV decoders receive only the hidden state as input, and not the SIV signal. \textbf{Only SIV Input}: The baseline model is used, but with the modification that the SIV is input to the decoder directly, as in our proposed approach.

\textbf{Noise and Missingness.} We assume that SIV signals are noise free. Here, we evaluate our model's performance as this assumption is relaxed. This is possible in our simulated dataset because we have ground truth carbohydrate values. In real data, not only may the magnitude of these values be inaccurate (they are estimated by the individual), but individuals may skip recordings altogether. In order to examine how unreliable carbohydrate values impact our approach, we randomly hide between 10\% and 50\% of the carbohydrate values (by setting their value to zero), and also add between 10\% and 50\% magnitude uniform random noise to the carbohydrate measurements (both after data generation, and to both training and testing data), to evaluate the performance of our model vs the stand-alone encoder/decoder baseline as carbohydrate values become unreliable. We report average forecast error across all individuals and 5 random noise-generation seeds for this analysis. We also evaluate our approach on a real dataset which is expected to have some degree of missingness and noise for comparison.

\section{Results and Discussion}

 We aim to answer the following questions.

    • Does our model offer improvement over baseline approaches in terms of forecast error and SIV Usage? 
    
    • Across individuals, when does our model offer the greatest improvement? 
    
    • What elements of our model improve performance? 
    
    • How do inaccurately measured SIVs impact our model?

\begin{figure}[t]
  \centering 
\hspace*{-.3cm}  \includegraphics[height=1.7 in]{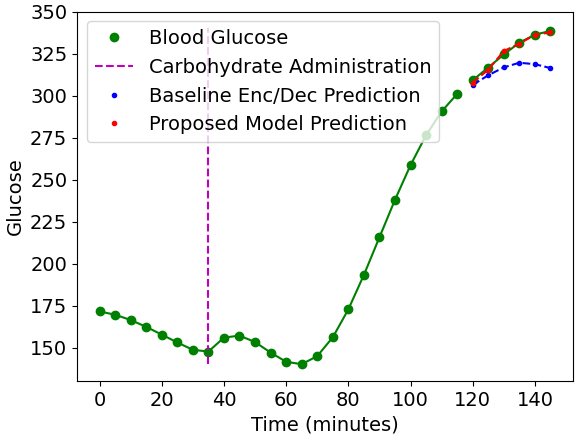}
\caption{A sample prediction for a simulated individual (adult\#004). Our model better accounts for the steep rise in the BG signal following a meal.}\label{fig:sampplot}
\end{figure}

\textbf{Improvement Over Baselines.} Our model outperforms baselines leading to lower average rMSE/ MAE and greater relative SIV usage (\textbf{Table \ref{main}}). Specifically, our model appears to account for the effect of the SIV on the target signal (\textit{e.g.}, accurately predicting sharp rises that the baseline encoder/decoder cannot account for: \textbf{Figure \ref{fig:sampplot}}). Naive SIV resampling approaches are generally outperformed by other baseline approaches, or perform similarly. Our proposed approach shows a large improvement over even the strongest baseline (rMSE 13.07 vs 14.14, paired t-test across individuals [t statistic/p-value]: 2.0/0.05).  We also examine the clinical benefit of forecasting with our approach using a Clarke error grid in \textbf{Appendix C}.

\begin{table}[htbp]
  \centering

\setlength{\tabcolsep}{2pt}
\hspace*{-0cm}\scalebox{0.9}{\begin{tabular}{lcc}

Model & rMSE [95\%CI](Usage)&MAE [95\%CI](Usage)\\
\hline
\\[-6pt]

 \textbf{BASELINES}\\
  Enc/Dec & 15.63[14.1,16.9](11.13) & 12.42[11.1,13.6]\hspace{1.5mm}(6.63)\\  
 SIV Fine-tune & 27.30[24.7,29.1](-0.54) & 22.22[19.9,23.9](-3.17)\\    
 SIV Initialize & 15.37[13.6,16.9](11.39) & 11.99[10.7,13.2]\hspace{1.5mm}(7.06)\\  
\vspace{2mm}
 Full Capacity & 14.14[12.7,15.3](12.62) & 11.21\hspace{1.5mm}[9.9,12.2]\hspace{1.5mm}(7.85)\\
\vspace{2mm}
 \textbf{PROPOSED} & 13.07[11.8,14.2](13.69) & 10.45\hspace{1.5mm}[9.4,11.4]\hspace{1.5mm}(8.61)\\

\textbf{ABLATIONS}\\
 No Gating & 13.93[12.6,15.0](12.84) & 11.11\hspace{1.5mm}[9.9,12.1]\hspace{1.5mm}(7.95)\\
  No Restriction & 13.12[11.8,14.2](13.64) & 10.43\hspace{1.5mm}[9.3,11.4]\hspace{1.5mm}(8.62)\\
 No SIV Input & 14.20[12.7,15.4](12.56) & 11.18\hspace{1.5mm}[9.9,12.2]\hspace{1.5mm}(7.87)\\
 Only  SIV Input & 13.97[12.6,15.1](12.79) & 11.12\hspace{1.5mm}[9.6,12.2]\hspace{1.5mm}(7.94)\\

\end{tabular}}
\caption{Mean forecasting error and SIV usage. Outcomes are reported as: Error [95\% confidence interval] (SIV Usage). Our proposed approach outperforms baselines and ablations. CIs were calculated using 1,000 bootstrap samples.}\label{main}
  \label{syn1} \label{mode1synX}
\end{table}

\textbf{Individual Level Results.} In the simulated dataset, we consider ten individuals who differ in terms of simulated physiological parameters. Here, we investigate how model performance with respect to the baseline Encoder/Decoder varies across these individuals. In particular, we identify settings in which our approach is most beneficial.

\begin{figure}[t]
 \centering 
 \hspace*{-.15cm} \includegraphics[height=1.7 in]{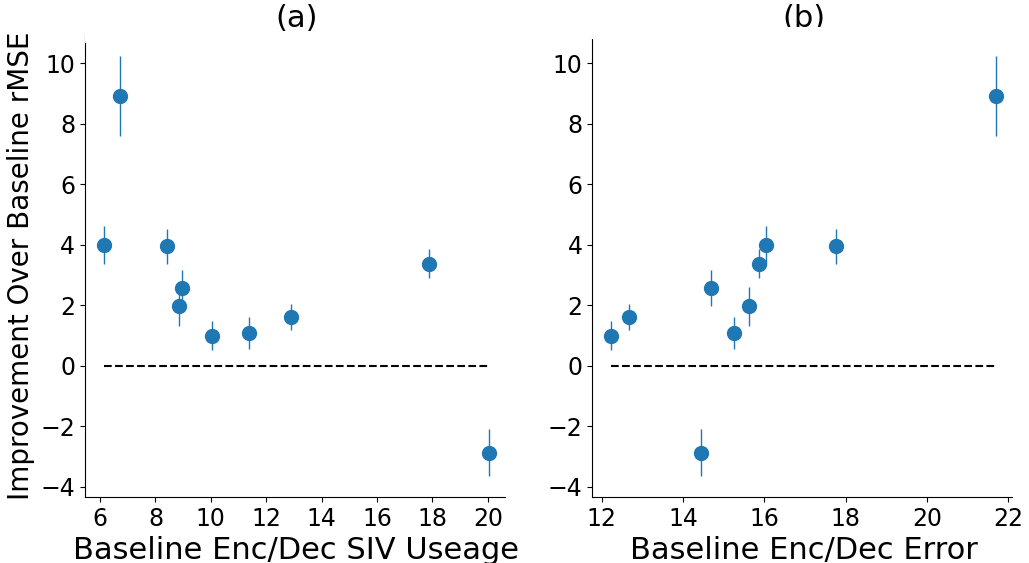}
\caption{ (a) Our architecture's improvement over baseline increases as baseline SIV usage decreases. (b) Improvement over baseline is positively correlated with baseline rMSE. Each point represents an individual in the simulated dataset, and errors bars represent standard error \textit{i.e.}, the standard deviation of 100 bootstrapped samples.}\label{fig:useerr}
\end{figure}

Our model's benefit over baseline varies inversely with the extent to which the baseline approach relies on the SIV  (\textit{i.e.}, SIV usage) across individuals (Pearson r=-0.65, p=0.041, \textbf{Figure \ref{fig:useerr} (a)}). This supports the hypothesis that our model's improved performance over the baseline is due in part to the increased usage of the SIV. For individuals for whom the baseline model was able to achieve high usage, our model was not necessary, but individuals with low baseline usage stood to benefit. We also observe a strong correlation between baseline forecast error and our approach's improvement  (r=0.80, p= 0.0056, \textbf{Figure \ref{fig:useerr} (b)}). This suggests that our approach addresses the deficits of the baseline at the individual level. There is higher rMSE variability across individuals for the baseline compared to the proposed approach (range: 9.5 vs 6.3), perhaps partially due to difficulties in SIV modeling, for which our model compensates.

\textbf{Ablations.}  Ablation analyses reveal that, in general, our approach's strong SIV usage and forecast accuracy are a result of the combined effects of each implementation detail, rather than any one component.  \textbf{Table \ref{main}} shows the results of our ablation study: removing any component results in a decrease in performance accuracy and SIV usage. Notably, removing the domain-guided restriction (\textit{i.e.,} the ReLU functions) results in the smallest effect on performance. This is likely because, in our simulated dataset, the effect of the insulin boluses and carbohydrate administrations are strong enough that the model can learn them easily without supervision. When the restriction element is included, it seems to be utilized by the model: randomly switching the sign of the ReLU component at test time in a post-hoc analysis drastically hampers performance, more than doubling rMSE. We expect this component to have a greater effect in situations where the impact of the SIV is more subtle.

\begin{figure}[t]
  \centering 
 \includegraphics[height=1.8 in]{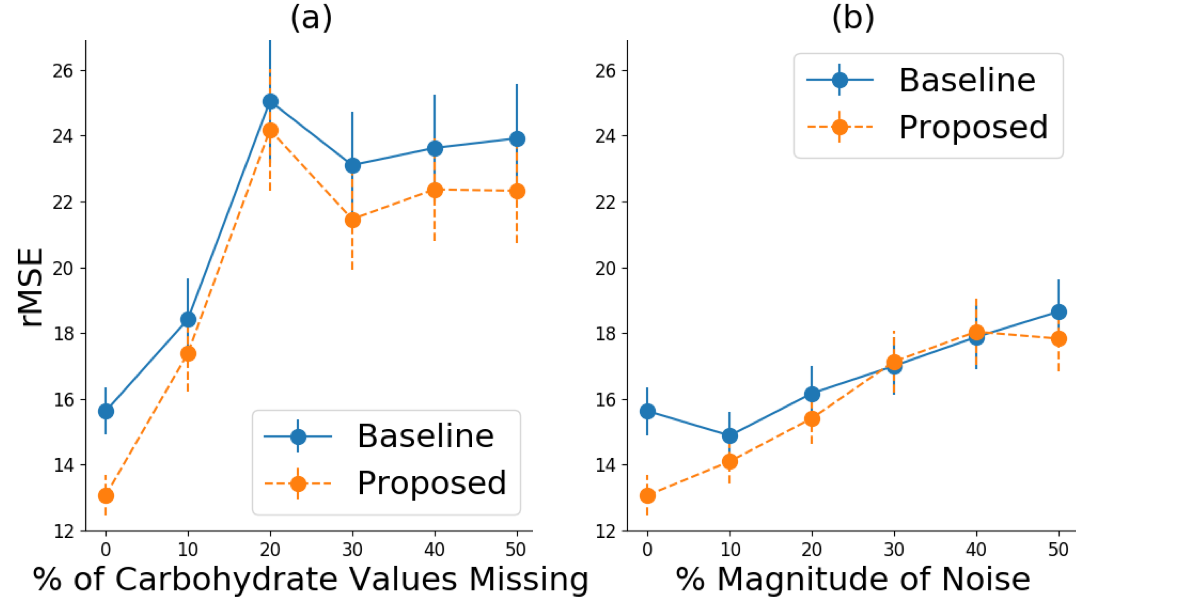}
\caption{Mean model performance across all individuals as (a) simulated carbohydrate values are hidden and (b) noise is added to carbohydrates. As noise and missingness increase, our model fails to reliably outperform the baseline. Errors bars represent standard error \textit{i.e.}, the standard deviation of 100 bootstrapped samples.  }\label{fig:carbmiss}
\end{figure}

\textbf{Noise and Missingness.} In the above experiments, we assume that the SIV is accurately measured. To quantify the impact of measurement noise on our approach, we perturb the SIV as described in the experimental setup section. We find that as missingness and noise increase, our approach's performance degrades (\textbf{Figure \ref{fig:carbmiss}}). Relative to the baseline, our approach is more impacted by corrupted SIV values, in part because of the increased dependence on the SIV (\textit{i.e.}, greater SIV usage). As expected, completely omitting carbohydrates has a greater effect than simply corrupting the magnitude.  Though performance decreases with increased noise/missingness, we are encouraged by the fact that our proposed approach remains competitive with the baseline.
 
 We further explore the effects of a corrupted SIV signal using the `Ohio' data, a real dataset generated by individuals with type 1 diabetes. While we cannot measure the amount of noise in the Ohio dataset, we expect it to be more in line with the simulated-noise setting than the noise-free setting.  Somewhat reassuringly, even in this noisy setting, our approach performs no worse than existing approaches and even provides a small benefit over baselines. Specifically, our approach consistently leads to lower forecast rMSE compared to all baselines, though performance gains are modest (rMSE=20.16 vs. the strongest baseline rMSE=20.36). Furthermore, in ablation analyses we found that the restriction component was beneficial for this dataset, supporting the hypothesis that domain knowledge insertion is beneficial for more challenging tasks (see \textbf{Appendix A} for full results).

\section{Conclusions}

The SIV problem arises in forecasting domains when the relative sparsity of an auxiliary signal makes learning its effect on a target signal challenging. We introduce the problem and propose a forecasting approach that leverages SIVs. Our approach isolates SIV dynamics and restricts them based on domain knowledge, achieving higher SIV-usage and stronger forecasting performance than baselines. While our approach assumes accurately measured SIVs, it performs no worse than baselines in the presence of missing or noisy SIV measurements. Though we focus on a specific use case for which we have a reliable simulator, we expect the SIV problem to arise frequently in healthcare. In such settings, SIVs are likely associated with time periods during which a patient is most vulnerable (\textit{i.e.}, medication administration). Therefore, prediction models that address the SIV problem could lead to more accurate predictions during time periods that are critical for health outcomes.

We are the first to identify the SIV problem that arises when using RNNs for multi-input forecasting and the first to propose a solution. While sparsely sampled variables (SSVs) have been studied \cite{missing}, the SIV problem is distinct. Interpolation approaches for addressing missingness and noise in SSVs are not directly applicable to the SIV setting. Although the SIV problem has not yet been addressed, several techniques have been proposed to learn inter-variable relationships in forecasting tasks, which in part inspire our approach. In the context of multi-input forecasting, Pantiskas et al. and Qin et al. use attention mechanisms to identify which variables to focus on \cite{attention1,cnnandatten}. However, in contrast to our approach, these approaches do not account for signals that are mostly zero-valued, nor do they incorporate domain knowledge. Beyond attention based mechanisms, in a probabilistic setting, normalizing flows have been used to directly model the joint probabilities between variables \cite{prob1,prob2}. However, SIVs are often too sparse to accurately estimate a joint probability. Several other approaches have been proposed to explicitly model inter-variable relationships \cite{cnn1,cnn2,cnnandatten,graphresid,sharedmulti}. However, none explicitly addresses the sparsity issue. Moreover, while some of these architectures separate the effects of variables, none use this isolation to restrict the effects based on domain knowledge as we do. There has been other work in forecasting that combines deep learning with domain knowledge to reduce the hypothesis space. However, researchers have relied on strong assumptions, \textit{e.g.}, structuring deep architectures to match clinical intuition \cite{gluexps}, combining deep approaches with physiological-model-based simulators \cite{wild}, and estimating expert judgements on model outputs via Monte-Carlo approximations \cite{domain}. In contrast, we only restrict the sign of the SIV network's hidden state, a more flexible approach.

While there are many different ways to forecast signals, we focus on RNN-based techniques. Our primary contributions are the identification of the SIV problem in forecasting and noting the failure of common RNN-based approaches when addressing this problem. We demonstrate how addressing the SIV problem can lead to improvements over directly comparable baselines. We do not claim SOTA in forecasting, but our findings could apply to many settings in which variants of RNNs are applied to forecasting problems with SIVs, which could include vital sign forecasting with medication administration as an SIV, stock prices with quarterly reports or news articles as SIVs, and  traffic forecasting with holidays or events as SIVs. Our approach is designed to utilize SIVs once an appropriate domain has been identified, but empirically identifying SIVs represents an interesting research direction. The two main ideas behind our approach include gating and output restriction. While neither of these methodological developments are unprecedented on their own, their combined application to the SIV problem poses a novel direction for forecasting in related domains.

\section*{Acknowledgements} This work was supported by JDRF (award no. 1-SRA-2019-824-SB).  The views and conclusions in this document are those of the authors and should not be interpreted as necessarily representing the official policies, either expressed or implied of JDRF.

\bibliography{aaai23}
\clearpage
\newpage

\appendix

\section{A Ohio Dataset Experiments}

\subsection{Data and Training Description}

This dataset includes both the OHIOT1DM 2018 and 2020 datasets, developed for the Knowledge Discovery in Healthcare Data Blood Glucose Level Predication Challenge \cite{kn:oc2}. The data pertain to 12 individuals, each with approximately 10,000 5-minute samples for training and 2,500 for testing, with carbohydrate administrations occurring every 88 timepoints on average, (median, [IQR]: 70, [56,134]), and insulin boluses occurring every 52 timepoints on average (36, [28,63]). 12\% of glucose values are missing, but we do not include windows with missing glucose values.

This dataset contains the same variables and is processed and analyzed identically to the simulated dataset, except as described here. For the real dataset we evaluated on the held-out test data from the challenges.  The remaining data were split into 80\% train and 20\% validation. Models were trained for at least 25 epochs, and then until validation data performance did not improve for 10 epochs.

The Ohio Dataset (Ohio T1D Blood Glucose Level Prediction Challenge, 2018 and 2020), can be made available through a data-use agreement with the owners: http://smarthealth.cs.ohio.edu/OhioT1DM-dataset.html.

\subsection{Primary Results}

On the Ohio dataset,  performance gains are more moderate (rMSE 20.16 vs 20.36, \textbf{Table \ref{mainreal}}), when compared to the simulated dataset. Multiple approaches exhibit negative SIV usage for the Ohio dataset, indicating that including the SIVs does more harm then good. We hypothesize that this is due to noise in the carbohydrate signal.

\begin{table}[htbp]

\setlength{\tabcolsep}{2pt}
\hspace*{-0cm}\scalebox{0.9}{\begin{tabular}{lcc}

Model & rMSE [95\%CI](Usage)&MAE [95\%CI](Usage)\\
\hline
\\[-6pt]

 Enc/Dec & 20.36[19.4,21.3](0.08) & 14.67[14.1,15.2](0.24)\\
 SIV Fine-tune & 21.74[20.9,22.6](-1.30) & 16.25[15.7,16.9(-1.35)\\
 SIV Initialize & 20.98[20.0,22.0](-0.54) & 14.99[14.4,15.6](-0.09)\\
Full Capacity & 20.98[20.0,21.9](-0.54) & 15.09[14.5,15.7](-0.18)\\
 Proposed & 20.16[19.3,21.1](0.28) & 14.64[14.1,15.2](0.27)\\

\end{tabular}}

  \centering 
  \caption{Forecasting Error and SIV usage for the real dataset. Outcomes are reported as: Error [95\% confidence interval] (SIV Usage). Our proposed approach outperforms baseline, although to a lesser degree than the simulated dataset. Confidence intervals were calculated from bootstraps with 1,000 resamples.}\label{mainreal}

  \label{syn1} \label{mode1synX}
\end{table}

\subsection{Individual Level Results and Ablations}

\begin{figure}[t]
 \centering 
 \hspace*{-.15cm} \includegraphics[height=1.8 in]{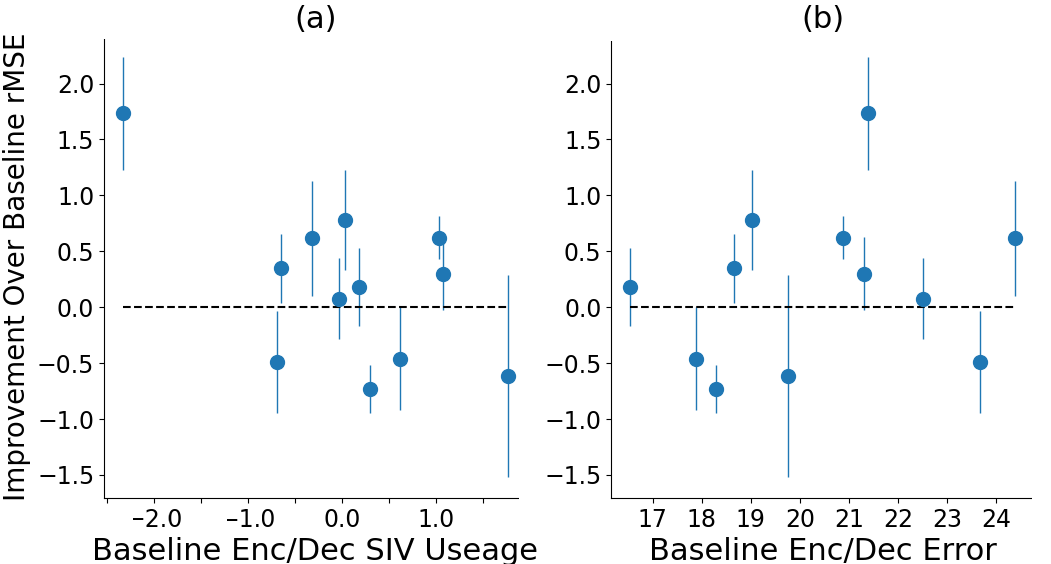}
\caption{(a) Our architecture's improvement over the encoder/decoder baseline vs baseline SIV usage for the Ohio dataset. Our method's benefit increases as baseline SIV usage decreases. (b) Improvement over baseline vs baseline prediction error for Ohio data, for each individual. Improvement over baseline is not correlated with baseline error.  }\label{fig:useerrR}
\end{figure}

With respect to the Ohio data, while the overall trend was the same as the simulated data, across individuals, the correlation between baseline error and our approach's improvement over baseline was not significant (r=0.22, p=0.49,  \textbf{Figure \ref{fig:useerrR} (b)}). We hypothesize that this again might be due to the presence of noise in the carbohydrate signal, which prohibits our model from accurately modeling the SIV signal (explored in Section 5.5). Alternatively, the intrinsic dynamics in the Ohio dataset may simply be more complex and thus result in more variability across individuals.  The association between baseline encoder/decoder SIV usage and improvement over baseline does hold for real data (\textbf{Figure \ref{fig:useerrR}} (a)), Pearson r=-0.59, p=0.042,

The restriction element is important for the Ohio dataset (\textbf{Table \ref{mode1ablR}}, rMSE increases to 20.38 from 20.16 when restriction is removed). This is likely because this dataset presents a more difficult challenge, compared to the simulated dataset, due to noise in the SIV signal and more complex target variable dynamics. For the Ohio dataset, we see a decrease in performance for each ablation. Our architecture works by isolating the effect that the SIV signal has on the target variable and enforcing consistency with domain knowledge. Although the domain knowledge is very general (we only restrict the signal direction), it improves performance, offering a benefit over isolation alone for the Ohio dataset. More restrictive model guidelines, such as directly restricting the architecture to use a detailed physiological model, could be beneficial, but during model development, we found that ``less is more,'' in that a small amount of restriction with significant flexibility was most effective. However, some sort of domain-knowledge-based-guidance is helpful to overcome the challenges posed by the SIV problem, since without it, it is difficult to learn anything useful from the small number of non-zero samples.

\begin{table}
  \centering

\setlength{\tabcolsep}{2pt}
\hspace*{0cm}\scalebox{0.9}{\begin{tabular}{lcc}

Model & rMSE [95\%CI](Usage)&MAE [95\%CI](Usage)\\
\hline
\\[-6pt]

& \multicolumn{2}{c}{Ohio}\\
 No Gating & 20.34[19.5,21.3](0.11) & 14.77[14.2,15.3](0.13)\\
 No Restriction & 20.38[19.5,21.3](0.06) & 14.73[14.2,15.3](0.18)\\
 No SIV Input & 20.71[19.8,21.6](-0.27) & 14.97[14.4,15.6](-0.07)\\
 Only  SIV Input & 20.52[19.6,21.5](-0.08) & 14.70[14.1,15.3](0.20)\\
 Proposed & 20.16[19.3,21.1](0.28) & 14.64[14.1,15.2](0.27)\\

\end{tabular}}
\caption{rMSE and MAE, with SIV usage, for each ablation. Outcomes are reported as: Error [95\% confidence interval] (SIV Usage). Confidence intervals were calculated from bootstraps with 1,000 resamples. }\label{syn1}
  \label{syn1} \label{mode1ablR}
\end{table}
\newpage
\section{B Impact of Carry-Forward Approach}

Utilizing the Carry-forward approach improves performance on both datasets for both the baseline encoder/decoder and our proposed approach (\textbf{Table \ref{CFTAB}}).

\begin{table}[htbp]
  \centering

\setlength{\tabcolsep}{2pt}
\hspace*{-.0cm}\scalebox{0.9}{\begin{tabular}{lcc}

Model & rMSE [95\%CI](Usage)&MAE [95\%CI](Usage)\\
\hline
\\[-6pt]
& \multicolumn{2}{c}{Simulated- Carry Forward}\\
 Enc/Dec & 15.63[14.1,16.9](11.13) & 12.42[11.1,13.6](6.63)\\  
 \textbf{Proposed} & \textbf{13.07[11.8,14.2](13.69)} & \textbf{10.45[9.4,11.4](8.61)}\\

& \multicolumn{2}{c}{Simulated- NO Carry Forward}\\
 Enc/Dec & 16.46[14.6,17.8](10.30) & 12.97[11.5,14.1](6.09)\\
 Proposed & 16.08[14.5,17.4](10.68) & 12.80[11.4,13.9](6.25)\\

&\multicolumn{2}{c}{Ohio- Carry Forward}\\
 Enc/Dec& 20.36[19.5,21.3](0.08) & 14.67[14.1,15.2](0.24)\\
 \textbf{Proposed} & \textbf{20.16[19.3,21.1](0.28)} & \textbf{14.64[14.1,15.2](0.27)}\\

&\multicolumn{2}{c}{Ohio- NO Carry Forward}\\
 Enc/Dec & 20.64[19.7,21.5](-0.20) & 14.98[14.4,15.6](-0.08)\\
 Proposed & 20.41[19.5,21.4](0.03) & 14.85[14.3,15.4](0.05)\\

\end{tabular}}

  \caption{Forecasting Error and SIV usage for both datasets, examining our primary baseline and proposed approach with and without our carry-forward approach. Outcomes are reported as: Error [95\% confidence interval] (SIV Usage). Both methods benefit from utilizing the carry-forward approach on both datasets. Confidence intervals were calculated from bootstraps with 1,000 resamples.}\label{CFTAB}
  
  \label{syn1} \label{CFTAB}
\end{table}

\begin{table}[htbp]
  \centering

\hspace*{-.5cm}\scalebox{0.8}{\begin{tabular}{lcc}

Region & Baseline&Proposed\\
\hline
\\[-6pt]
A & 97.1& 97.8 \\
B & 2.36 & 1.43\\
C & 0.00& 0.00\\
D & 0.05& 0.08\\
E & 0.00& 0.00\\
\end{tabular}}
\caption{Proportion of points in each region of the Clarke Error Grid. Region A represents strong forecasts, while regions C through E represent potentially catastrophic errors}\label{CEGT}
  \label{syn1} \label{CEGT}
\end{table}

\section{C Clarke Error Grid}

\begin{figure}[t]
 \centering 
 \hspace*{-.1cm} \includegraphics[height=2.3 in]{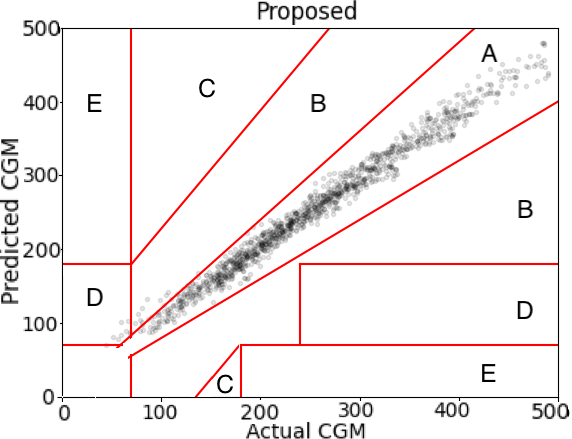}
\caption{ Clarke error grid for the proposed approach.}\label{fig:ceg1}
\end{figure}

\begin{figure}[t]
 \centering 
 \hspace*{-.1cm} \includegraphics[height=2.3 in]{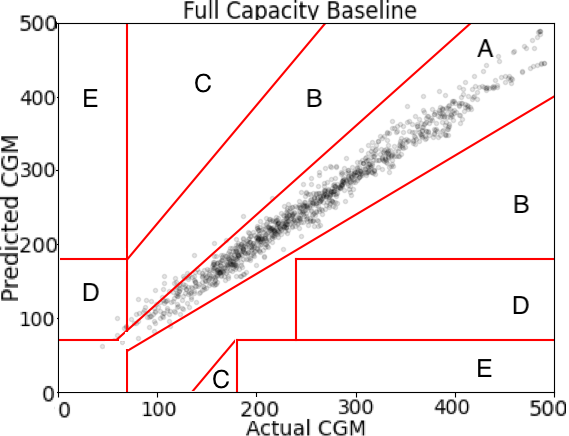}
\caption{Clarke error grid for the full capacity baseline. }\label{fig:ceg2}
\end{figure}

A Clarke error grid demonstrates where a forecaster could lead to catastrophic failure; predicted BG values are compared to true values, and regions where making treatment decisions based on forecasts would lead to poor health outcomes are highlighted. Clarke error grids for our approach and the best performing baseline on the simulated dataset, are shown in \textbf{Figure \ref{fig:ceg1}} and \textbf{Figure \ref{fig:ceg2}}, respectively. Both approaches demonstrate fairly strong performance, but our approach has 98\% of points in region A, while the baseline has 97\% of points in region A (\textbf{Table \ref{CEGT}}). Region A represents the region where utilizing the forecasts for BG control would reliably lead to good health outcomes. While one percentage point is a modest improvement, every prediction is important in a clinical setting. This illustrates that while our approach is not a complete solution to reliable BG forecasting, it a step in the right direction.

\end{document}